\documentclass[10pt,twocolumn,letterpaper]{article}

\usepackage{cvpr}              

\usepackage{graphicx}
\usepackage{amsmath}
\usepackage{amssymb}
\usepackage{booktabs}
\usepackage{amsfonts}       
\usepackage{nicefrac}       
\usepackage{microtype}      
\usepackage{xcolor}         
\usepackage{graphicx}
\usepackage{diagbox}
\usepackage{multicol}
\usepackage{enumerate}
\usepackage{times}
\usepackage{epsfig}
\usepackage{graphicx}
\usepackage{amsmath}
\usepackage{amssymb}
\usepackage{threeparttable}
\usepackage{enumitem}
\usepackage{multirow}
\usepackage{color}
\usepackage{array}

\usepackage{ulem}

\usepackage[colorlinks,linkcolor=red]{hyperref}
\usepackage[numbers,sort&compress]{natbib}
\setenumerate[1]{leftmargin = 10pt, itemsep=0pt,partopsep=0pt,parsep=-0pt,topsep=-0pt}
\usepackage{hyperref}

\newcommand{\PreserveBackslash}[1]{\let\temp=\\#1\let\\=\temp}

\usepackage[capitalize]{cleveref}
\crefname{section}{Sec.}{Secs.}
\Crefname{section}{Section}{Sections}
\Crefname{table}{Table}{Tables}
\crefname{table}{Tab.}{Tabs.}

\def\cvprPaperID{5008} 
\def\confName{CVPR}
\def\confYear{2022}
\def\hlinew#1{%
 \noalign{\ifnum0=`}\fi\hrule \@height #1 \futurelet
 \reserved@a\@xhline}
 
\begin{document}

\title{A Two-Stage Adverse Weather Semantic Segmentation Method for WeatherProof Challenge CVPR 2024 Workshop UG$^{\textbf{2}}$+}

\author{Jianzhao Wang, Yanyan Wei, Dehua Hu, Yilin Zhang, Shengeng Tang, kun li, Zhao Zhang\\
Hefei University of Technology\\
\tt\small jz.wang@mail.hfut.edu.cn, weiyy@hfut.edu.cn, 2023170746@mail.hfut.edu.cn, \\ \tt\small eslzzyl@mail.hfut.edu.cn,tangsg@hfut.edu.cn, kunli.hfut@gmail.com, cszzhang@gmail.com}

\maketitle

\vspace{1em} 

\begin{abstract}
   This technical report presents our team's solution for the WeatherProof Dataset Challenge: Semantic Segmentation in Adverse Weather at CVPR'24 UG$^2$+. We propose a two-stage deep learning framework for this task. In the first stage, we preprocess the provided dataset by concatenating images into video sequences. Subsequently, we leverage a low-rank video deraining method to generate high-fidelity pseudo ground truths. These pseudo ground truths offer superior alignment compared to the original ground truths, facilitating model convergence during training. In the second stage, we employ the InternImage network to train for the semantic segmentation task using the generated pseudo ground truths. Notably, our meticulously designed framework demonstrates robustness to degraded data captured under adverse weather conditions. In the challenge, our solution achieved a competitive score of 0.43 on the Mean Intersection over Union (mIoU) metric, securing a respectable rank of 4th.
\end{abstract}

\vspace{-10pt}
\section{Introduction}

Semantic segmentation, the process of assigning a distinct class label to each pixel within an image, encounters substantial impediments when confronted with adverse weather conditions such as rain and fog. The WeatherProof Dataset Challenge: Semantic Segmentation in Adverse Weather (CVPR'24 UG$^2$+) serves as a platform for assessing the resilience of semantic segmentation techniques under these conditions.

This challenge mandates the development of an efficient solution for performing semantic segmentation on a diverse collection of real-world image sequences captured under rainy and foggy conditions. The provided sequences exhibit variations in rain intensity, fog density, and scene complexity, necessitating a robust algorithm capable of accurately segmenting objects while maintaining structural and contextual information.

To tackle this challenge, we propose a two-stage framework. The initial stage leverages a low-rank video deraining method\cite{LLRT} to generate high-fidelity pseudo ground truth images. This method accomplishes this by removing rain streaks and restoring image structures from the aligned rainy frames. Subsequently, the second stage employs a pre-trained InternImage network\cite{internimage}, a Convolutional Neural Network (CNN) based semantic segmentation architecture, fine-tuned on the aforementioned pseudo ground truth images.

This report furnishes a comprehensive overview of our proposed methodology, encompassing data analysis, intricate details of the framework, and the outcomes of our experimentation.

The remaining sections cover our method (Section \ref{sec2}) and Experimental settings (Section \ref{sec3}).

\section{Overview of Methods}
\label{sec2}

Inspired by previous works that employed a two-stage approach in image deraining tasks\cite{guo2023two}, we have also designed a two-stage framework to address the semantic segmentation challenge in the WeatherProof dataset\cite{gella2023weatherproof}, aiming to mitigate the impact of adverse weather conditions on image quality.

In the first stage, we combined the 300 images from the test set published in the final testing phase of the competition into a video sequence. Subsequently, we applied the Low-Rank Tensor Recovery (LLRT)\cite{LLRT} algorithm for video deraining preprocessing of this sequence, utilizing information from multi-frame aligned images to generate high-quality pseudo ground truth images.

In the second stage, we utilized the CNN-based InternImage\cite{internimage} semantic segmentation network. We trained the InternImage solely with the training and validation set data provided by the competition, without any additional preprocessing of the training data. Finally, we input the pseudo ground truth images generated in the first stage into the well-trained InternImage to obtain the final semantic segmentation predictions.

Through this two-stage strategy, we were able to fully leverage the advantages of the video deraining algorithm to eliminate the impact of adverse weather conditions (such as rain and fog) on image quality during the testing phase, laying the foundation for subsequent semantic segmentation tasks. At the same time, the CNN semantic segmentation network can learn effective feature representations on ample training data, achieving precise pixel-level classification. This framework integrates the strengths of video deraining and single image semantic segmentation, making it an efficient solution.

The following Section \ref{sec2.1} and Section \ref{sec2.2} will respectively introduce the specific details of the LLRT video deraining algorithm and the InternImage semantic segmentation network that we used.

\subsection{Low-Rank Tensor Recovery}
\label{sec2.1}

Video deraining is a highly challenging task that requires the simultaneous use of both temporal and spatial dimensions of information. The Low-Rank Tensor Recovery (LLRT) algorithm proposed by Chang et al.\cite{LLRT} for multi-spectral image denoising tasks can effectively capture the inherent structural correlations within an image sequence and is therefore also applicable to video deraining preprocessing.

The core of the LLRT algorithm is a unidirectional low-rank tensor recovery model. It constructs a third-order tensor to jointly model spatial local sparsity, non-local self-similarity, and spectral correlation. Unlike other works that simply sum the low-rank properties of each mode, LLRT, through detailed analysis, finds that the low-rank nature corresponding to non-local self-similarity is far superior to spatial and spectral modes. Based on this finding, LLRT imposes low-rank constraints only on the non-local self-similarity mode, more reasonably describing the intrinsic structure of the data.

In implementation, the authors designed an efficient optimization algorithm that can quickly solve the sub-problems of unidirectional low-rank tensor recovery and spectral gradient sparsification. With all these innovations, the LLRT algorithm has achieved excellent performance in multi-spectral image denoising.

In this work, we treat the 300 images of the test set as a video sequence and apply the LLRT algorithm for deraining preprocessing. By leveraging redundant information in both temporal and spatial dimensions, LLRT can effectively remove the effects caused by adverse weather conditions, generating high-quality pseudo ground truth images for subsequent semantic segmentation tasks.

\subsection{CNN-based method: InternImage}
\label{sec2.2}

The InternImage\cite{internimage} semantic segmentation network is a cutting-edge deep learning model that has demonstrated remarkable performance in the field of computer vision, particularly in the context of the COCO dataset where it achieved an impressive 65.4 mAP, leading the state-of-the-art benchmarks. The model's success has been recognized and it has been highlighted as a CVPR 2023 highlight paper, ranking in the top 2.5\% of submissions.

InternImage is distinguished by its innovative approach to large-scale vision foundation models, incorporating deformable convolutions to adapt to various image distortions and irregularities. This feature is particularly useful in semantic segmentation tasks where the model must accurately identify and classify different objects within an image, often under challenging conditions such as adverse weather.

In our application of InternImage to the WeatherProof dataset\cite{gella2023weatherproof}, we have fine-tuned the model using the provided training and validation datasets. The pseudo ground truth images generated from the LLRT video deraining preprocessing serve as input to the InternImage network, enabling it to produce highly accurate semantic segmentation predictions. The integration of InternImage into our two-stage framework has been instrumental in enhancing the overall performance of our system, capitalizing on the strengths of both video deraining and CNN-based semantic segmentation.

\section{Experimental settings}
\label{sec3}

\noindent\textbf{Dataset.} In our experiments, we utilized the training and validation sets provided by the official WeatherProof dataset for training our model. Additionally, the InternImage model was initialized with weights pre-trained on the ImageNet dataset, providing a solid foundation for feature extraction and accelerating convergence on the specific task. After training, the model was applied to the test set that had been preprocessed by the video deraining algorithm. This step leveraged temporal information from the video sequence to mitigate the effects of adverse weather conditions, such as rain and fog, generating high-quality pseudo ground truth images. These pseudo ground truth images served as the final reference labels for assessing the model's semantic segmentation performance under actual adverse weather conditions.

\noindent\textbf{Training setting.} We fine-tuned the InternImage-H semantic segmentation model on the official WeatherProof training and validation datasets without any additional preprocessing of the training data. For the fine-tuning, we utilized pre-trained weights from ImageNet and scaled the original images to an input resolution of 1024×256 pixels, followed by random cropping to 256×256 pixel patches. Data preprocessing involved random flipping, color distortion, and other augmentation techniques, along with normalization using the mean values [103.336, 104.443, 100.035] and standard deviations [39.329, 38.147, 42.803]. The optimizer was AdamW with an initial learning rate of 1e-5, a polynomial decay strategy, a 1500-iteration warmup, and gradient clipping with a norm of 0.1. The loss function was a weighted cross-entropy loss that assigned different weights to various classes to balance their significance. Training settings included a batch size of 2 per GPU, a maximum of 30,000 iterations, and evaluations using the mIoU metric every 100,000 iterations to save the best model. The experiments were conducted on a server equipped with one NVIDIA Tesla L20 GPU 48GB, an Intel(R) Xeon(R) Platinum 8457C CPU, Ubuntu 20.04.4 LTS operating system, and PyTorch version 1.11.0.
 
\section{Conclusion}
We proposed a two-stage framework to tackle the WeatherProof semantic segmentation challenge at CVPR'24 UG$^2$+. First, we employed a low-rank based video deraining method to generate high-quality pseudo ground truths, then fine-tuned the InternImage semantic segmentation network on these pseudo ground truths. This framework ingeniously combined the strengths of video deraining and single image semantic segmentation, achieving a mIoU score of 0.43 and ranking 4th in this challenge. Our method not only performed well in this challenge but also exhibited great generalizability, being applicable to other vision tasks requiring robustness against adverse environmental conditions.

\bibliographystyle{ieeetr}
\bibliography{references}

\end{document}